# Mining Themes in Clinical Notes to Identify Phenotypes and to Predict Length of Stay in Patients admitted with Heart Failure

Ankita Agarwal
Dept. of Computer Sci. and Engr.
Wright State University
Dayton, U.S.A.
agarwal.15@wright.edu

Tanvi Banerjee
Dept. of Computer Sci. and Engr.
Wright State University
Dayton, U.S.A.
tanvi.banerjee@wright.edu

William L. Romine
Dept. of Biological Sciences
Wright State University
Dayton, U.S.A.
william.romine@wright.edu

Krishnaprasad Thirunarayan
Dept. of Computer Sci. and Engr.
Wright State University
Dayton, U.S.A.
t.k.prasad@wright.edu

Lingwei Chen
Dept. of Computer Sci. and Engr.
Wright State University
Dayton, U.S.A.
lingwei.chen@wright.edu

Mia Cajita
College of Nursing
University of Illinois Chicago
Chicago, U.S.A.
mcajit2@uic.edu

*Abstract*— Heart failure is a syndrome which occurs when the heart is not able to pump blood and oxygen to support other organs in the body. Treatment and management of heart failure in patients include understanding the diagnostic codes and procedure reports of these patients during their hospitalization. Identifying the underlying themes in these diagnostic codes and procedure reports could reveal the clinical phenotypes associated with heart failure. These themes could also help clinicians to predict length of stay in the patients using their clinical notes. Understanding clinical phenotypes on the basis of these themes is important to group patients based on their similar characteristics which could also help in predicting patient outcomes like length of stay. These clinical phenotypes usually have a probabilistic latent structure and hence, as there has been no previous work on identifying phenotypes in clinical notes of heart failure patients using a probabilistic framework and to predict length of stay of these patients using data-driven artificial intelligence-based methods, we apply natural language processing technique, topic modeling, to identify the themes present in diagnostic codes and in procedure reports of 1,200 patients admitted for heart failure at the University of Illinois Hospital and Health Sciences System (UI Health). Topic modeling identified twelve themes each in diagnostic codes and procedure reports. These themes revealed information about different phenotypes related to various perspectives about heart failure, which could help to study patients' profiles and discover new relationships among medical concepts. Each theme had a set of keywords and each clinical note was labeled with two themes — one corresponding to its diagnostic code and the other corresponding to its procedure reports along with their percentage contribution. We used these themes and their percentage contribution to predict length of stay. We found that the themes discovered in diagnostic codes and procedure reports using topic modeling together were able to predict length of stay of the patients with an accuracy of 61.1% and an Area under the Receiver Operating Characteristic Curve (ROC AUC) value of 0.828.

Research reported in this publication was supported by National Institutes of Health award number R01AT010413NCCIH. Any opinions, findings, and conclusions or recommendations expressed in this material are those of the authors and do not necessarily reflect the views of the NIH. The experimental procedures involving human subjects described in this paper were approved by the Institutional Review Board.
Corresponding author: Ankita Agarwal (agarwal.15@wright.edu)

## I. INTRODUCTION

Heart failure (HF) syndrome occurs in a situation when the heart cannot fill up with enough blood or when the heart is too weak to pump properly to support other organs in the body as well as it should for the body's needs [1]. It is a condition which progresses through different stages ranging from "high risk of developing heart failure" to "advanced heart failure" [1]. Heart failure can be sudden due to a medical condition or injury but mostly it develops slowly due to long-term medical conditions [2]. Some conditions that could cause heart failure are arrhythmia, cardiomyopathy, coronary artery disease, congenital heart disease, diabetes, obesity, and hypertension [2]. Admission of patients in hospitals for heart failure incurs costs to the patients and increases the burden on hospitals for resource allocation [3] which could be reduced by preventing hospitalization or reducing the length of stay of patients [3]. Length of stay is defined as the time between hospital admission to discharge of a patient. In the United States alone, over 6.2 million people have heart failure which costs about 31 billion dollars to the US healthcare system [4]. Therefore, predicting length of stay in patients admitted for heart failure is important.

Electronic health records (EHRs) of a patient store the systematized longitudinal collection of patient's health information electronically [5]. EHRs can be used to build predictive models and effective clinical decision support systems by leveraging artificial intelligence techniques like machine learning and natural language processing [6]. A typical EHR contains diagnostic codes, procedure reports, discharge summaries, vitals, medications administered, laboratory tests, and imaging reports collected during a patient's visit to a healthcare facility. Identifying and studying the abstraction of the underlying medical concepts in these clinical health records help us to identify the clinical phenotypes related to a disease and to determine appropriate discharge destination and care for the patients. These clinical phenotypes have a latent structure and so earlier researchers have used probabilistic methods like topic models [7], probabilistic graphical model [8], and latent class analysis [9] to study clinical concepts associated with the disease. Further, [7] studied the importance of thematic structure of the notes to clinicians in inferring the hidden relationships and phenotypes.

Length of stay of a patient also depends on diagnosis [10] and procedure reports [11] of a patient. The implication of procedure reports and diagnostic codes for supporting to predict length of stay has been studied earlier [11], [12], which, however, still suffers from a limitation that these implications have not been studied in predicting length of stay in patients admitted for heart failure. As such, in this study, we study the underlying themes in the clinical notes from diagnostic codes and procedure reports to characterize patients admitted for heart failure to predict their length of stay.

To the best of our knowledge, there has been no previous work on identifying the phenotypes of heart failure patients by modeling their disease in latent space and uncovering the underlying themes within their clinical notes. Additionally, earlier researchers have not implemented data-driven artificial intelligence-based methods to predict length of stay using clinical notes of patients admitted for heart failure. So, in this study, we proposed topic modeling to study the underlying medical concepts in the form of themes to uncover the clinical phenotypes in the patients' records admitted for heart failure. Additionally, these themes could be used to predict their length of stay in these patients. Our rationale is that by using the natural topics detected by the algorithm, we are more likely to create a generalizable solution that can work on data collected in the future. Thus, we explored the following research questions in this study:

1. Can we identify the clinical phenotypes of the patients admitted for heart failure in the form of themes?
2. Can these themes be used to predict length of stay of a patient?

## II. RELATED WORK

Natural language processing methods like topic modeling is a probabilistic framework to identify the themes underlying the dataset. Previous studies have shown that the thematic structure of topic modeling offers a good approach for interpreting clinical notes by the medical professionals and assisting them in making clinical decisions [13]. In [14], the researchers found that the probabilistic phenotyping of the disease help to identify the sub type and characteristics of the underlying disease which allows for the probabilistic reasoning to clinicians. [15] developed a visualization tool based on topic models on clinical notes to characterize disease sub-types in a patient on the basis of the probabilistic phenotypes present in them. They used a probabilistic framework to explain the latent disease processes and also the temporal evolution of the prevalence of phenotypes in a disease subtype by comparing their probabilities. They validated the result of the topic models (that captures patterns and comorbid phenotypes) with the help of a developmental behavioral pediatrician.

Additionally, topic modeling has been used to improve predictive modeling of HIV diagnosis [16] using unstructured texts in clinical notes. Similarly, it has been used to understand dementia and its health care trajectories by analyzing clinical notes [17] and to predict psychiatric hospital readmissions [18]. In [19], researchers used topic modeling to gain insights into cancer from clinical notes and to study the relationships between patient clinical notes and their underlying genetics. [20] discovered associations among diagnostic codes using topic modeling. They found that disease groups based on topic modeling have statistical significance and can reveal semantic commonalities among diseases. In our previous work [21], we studied that topic models could uncover themes in the procedure reports which can help us to understand the common findings in the heart failure patients.

The importance of predicting patients' length of stay has been studied by earlier researchers who determined that length of stay is associated with patient satisfaction in emergency care [22] and in hospital economic performance [23]. [24] created a decision support system to predict length of stay using a dataset of hospitalized patients and a two-level classification system. They used demographic, clinical (diagnostic codes), geographical and administrative factors to build their decision support system. They classified the patients in the dataset according to the number of days in the hospital such as 1, 2, 3, 4, 5, and 5+ days. On the first level of the classification scheme, they used a classifier to predict length of stay of patients between "1–2" days, "3–5" days, or "5+" days and used a second-level classifier to derive a more specialized decision. They used naive Bayes, multilayer perceptron, sequential minimal optimization (SMO), k-nearest neighbors (kNN), and random forests (RF) as the classification algorithms and achieved an accuracy of 41.96%, 51.31%, 67.67%, 72.34%, and 67.59%, respectively, for one-level classifier. For a two-level classification, they achieved an accuracy of 78.53% using a combination of RF as level one classifier and kNN and RF as level two classifier.

In a different study, these researchers used the same dataset and features to predict length of stay in patients using self-supervised algorithms [25]. They proposed that self-supervised algorithms could achieve good accuracy in predicting length of stay of patients where there is little labeled data and much unlabeled data. They evaluated the performance of three self-supervised algorithms - self-training, co-training, and tri-training. They evaluated them by deploying naive Bayes (NB), multilayer perceptron (MLP), sequential minimum optimization, kNN algorithm, C4.5 decision tree algorithm, and PART as base learners and were able to achieve an accuracy of 62% to 64% using these algorithms to predict length of stay. [26] used data mining techniques to predict length of stay of cardiac patients. They used data from patients with coronary artery disease (CAD) and used decision trees, support vector machines (SVM), and artificial neural networks (ANN) as machine learning methods to predict length of stay. Their work was limited to patients with coronary artery disease and did not include data about patients with heart failure.

We address the gaps in the literature by modeling the clinical notes of heart failure patients in a probabilistic framework in the form of themes using topic modeling to identify phenotypes and to predict their length of stay during each hospitalization.

## III. METHODS

### A. Dataset Preparation

We obtained the dataset of 1,200 patients who were 18 years or older during their first hospitalization through the University of Illinois (UI) Chicago Center for Clinical and Translational Science (CCTS) Biomedical Informatics Core. It comprised the EHRs of patients admitted to UI Health between August 2016 to August 2021. A patient could have one or more records depending on the number of times a patient was hospitalized during this period. So, there were a total of 2,896 hospitalization records with admission and discharge dates of these patients. Each record was identified by a unique encounter id to de-identify the patient and their hospitalization. During a patient's hospitalization, diagnostic code (primary and secondary diagnostic codes) and information about the procedures performed on them was maintained. Primary diagnostic codes are provided at the time of inpatient encounter and they describe the diagnosis that were the most serious and/or resource-intensive during the hospitalization. In contrast, secondary diagnostic codes are those conditions that coexist at the time of admission, or develop subsequently, and that affect the patient care for a particular hospital stay.

A total of 161 records did not have diagnostic code information and so these records were discarded from our study. We combined the primary and secondary diagnosis codes together to get a list of diagnostic codes of a patient during a particular hospital stay. These diagnostic codes were coded using the International Classification of Diseases, Tenth Revision, Clinical Modification (ICD-10 CM) system. To identify the themes, present in these diagnostic codes, we first converted these codes to text. Agency for Healthcare Research and Quality (AHRQ) created a dataset known as Clinical Classifications Software Refined (CCSR) for ICD-10-CM Diagnoses, aggregating more than 70,000 ICD-10-CM diagnosis codes into over 530 clinically meaningful categories[1]. We used this dataset to convert the ICD-10 CM diagnosis codes to text. After converting the codes to text, we removed the word 'Heart Failure' from our further analysis as this word would not contribute to understanding meaningful themes in our dataset.

The procedure dataset contained information about the procedures performed on patients during and in between their hospitalization. Each procedure record was labeled with the encounter id like the diagnostic code dataset to identify the visit or hospitalization record during which the procedure was performed. A patient can have one or more procedure records during a hospitalization. There was a total of 15,966 clinical notes of 1,200 patients containing details of procedures performed on these patients along with the findings of these procedures and/or impressions or conclusions as given by the consulting doctor. An example of the clinical note with procedure record in our dataset is shown in Figure 1.

We extracted the impression or conclusion attribute from these notes. If the notes did not contain either of these attributes, we extracted the findings attribute from these notes for our analysis. We discarded the notes which did not contain either the impression, conclusion or findings attribute as they would not contribute to our textual analysis of the notes. Thus, we were left with a total of 15,634 procedure records. Thereafter we removed stop words and numbers from the procedure reports as these would not give any meaningful information.

PROCEDURE: CHEST X-RAY (PA, LATERAL).

FINDINGS: There is redemonstration of moderate cardiomegaly. There is pulmonary vascular redistribution with interstitial opacity. The aorta is tortuous with calcifications of the arch. No pleural effusion is visualized. There is no pneumothorax.

IMPRESSION: Redemonstration of moderate cardiomegaly with interstitial edema.

Fig. 1. An example of a procedure report in our dataset

*B. Negation Detection*

We performed negation detection on the procedure reports as it is a crucial part of processing clinical notes soundly. For example, an impression represented as 'The prostate gland is normal in size with no pelvic lymphadenopathy' means a normal finding, whereas, the impression represented as 'The prostate gland is enlarged with pelvic lymphadenopathy' would mean a disease or an abnormal finding in the patient. So, we employed a negation detection tool to preserve the semantics [27] of the procedure reports. Thus, after detecting the negated phrases, we explicitly affixed 'no' in front of that phrase. For example, the negated 'pleural effusions' was rendered 'no pleural effusions.'

*C. Named Entity Recognition*

As we were interested in predicting length of stay in a patient which could depend on the abnormalities or diseases being present or absent in a patient during a particular hospitalization, after performing negation detection on procedure reports, we retained the phrases containing the disease entity which were identified using the Python scispacy model[2]. So, in this way, phrases specific to the procedure performed, organ system, and other generic words in clinical notes like physician, electronically signed, personal name, alphanumeric id, and preliminary were removed as they would not be relevant to our analysis. As a patient can have one or more procedure reports during a particular visit, after performing this analysis, we combined all the procedure reports of the same patient during a particular hospital stay identified by the encounter id and removed duplicate phrases from the same patient's records during a particular stay.

Finally, we combined these procedure reports of a patient with their corresponding diagnostic codes text. So, we were left with a total of 2,486 records with procedure reports and diagnostic data of the patients for our analysis.

*D. Topic Modeling and Clinical Phenotyping*

We implemented two topic models, one on the diagnostic codes and the other on the procedure reports using a probabilistic framework known as Latent Dirichlet Algorithm (LDA). Topic models are used to discover the latent semantic structures in the documents statistically. LDA considers documents as a probability distribution of topics and each topic as a probability distribution over a collection of words. These topics and words have hyperparameters of the Dirichlet prior distributions known as alpha and beta respectively. In this method, first, each word in every document is assigned a random topic. Thereafter, the proportion of words in each document that are assigned to a particular topic and the proportion of assignments to a particular topic over all documents that come from this word is calculated. This process is reiterated until a steady state is reached and all assignments are acceptable.

Each topic in the diagnostic codes and procedure reports had a set of words known as keywords. We named each topic with a theme based on these keywords. These themes were studied to uncover the clinical phenotypes in the patients' records and to predict the length of stay of heart failure patients. To perform topic modeling on diagnostic codes, we converted the text into a vector representation using TF-IDF weighting and identified the optimal number of topics by generating a coherence plot using the Coherence Model [28].

---

[1] https://www.hcup-us.ahrq.gov/toolssoftware/ccsr/ccs_refined.jsp

[2] https://allenai.github.io/scispacy/

We found that 12 would be the optimal number of topics to explain the features in the diagnostic codes. Thereafter, we fitted a 12-topic Latent Dirichlet Algorithm (LDA) model and labeled each clinical note with the dominant topic. So, each note now had a topic number and percentage contribution of that topic. Then we identified the theme manually for each topic by looking at the keywords of each topic and representative notes which had the percent contribution of 0.80 or higher to the topic. Likewise, we performed the same analysis on the procedure reports. Similarly, for the procedure reports, we found 12 to be the optimal number of topics to explain the features in them.

To compare the performance of the fitted topic models in predicting length of stay on texts represented in TF-IDF format, with a baseline approach we also fitted 12-topic models on diagnostic codes and procedure reports by representing their texts in the bag-of-words (BOW) format.

*E. Length of Stay Prediction*

To predict the length of stay of patients, we calculated the number of days between the hospital admission and discharge date of the patient during a particular stay. Then based on the number of days, we categorized these into five categories: very short-term stay (0-1 day or 24 hours hospitalization), short-term stay (2-7 days), medium-term stay (8-14 days), long-term stay (15-21 days) or very long-term stay (more than 21 days). The number of records corresponding to very short-term stay, short-term stay, medium-term stay, long-term stay, and very long-term stay were 266, 1345, 559, 170, and 151 respectively. There was a maximum number of records for short-term stays. The histogram corresponding to the number of records belonging to each category of length of stay is shown in Figure 2.

We used topic numbers and percentage contribution of each theme for procedure records and diagnostic codes as predictor variables to predict the length of stay. We implemented three categories of machine learning classification algorithms, namely similarity based algorithm, linear classification models, and ensemble-based model using 70% training and 30% test data for this purpose using python scikit-learn library[3].

We implemented K-nearest neighbors (KNN) as a similarity-based algorithm, multinomial logistic regression (MLR) and support vector machines (SVM) as linear classification models and AdaBoost classifier (AC) and random forest (RF) as ensemble-based models. For KNN, we used the value of nearest neighbors as 3, for AC and RF, we used the number of estimators as 100. As there was an imbalance in the number of instances corresponding to each type of visit, we first used Synthetic Minority Over-sampling Technique (SMOTE) to balance the data [29]. The results of predicting length of stay using different machine learning algorithms are shown in Table 3.

The overall methodology to identify clinical phenotypes and predict length of stay in patients admitted for heart failure is shown in Figure 3.

---

[3] https://scikit-learn.org/stable/

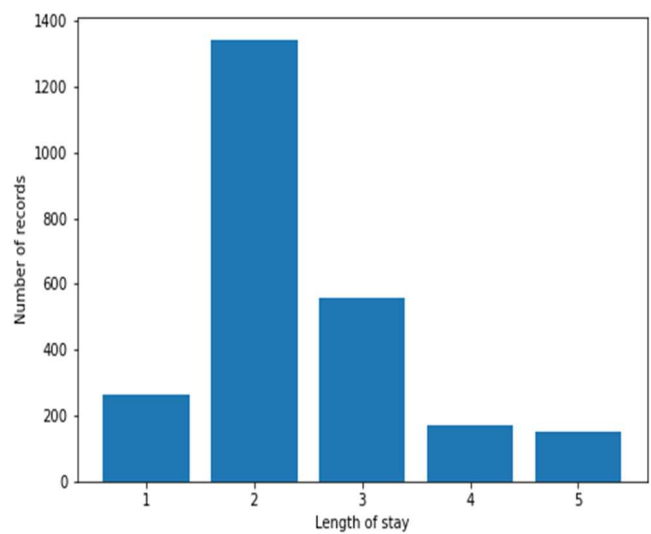

Fig. 2. Distribution of number of records for each type of length of stay

## IV. RESULTS

*A. Identifying the Clinical Phenotypes in the Form of Themes*

After implementing LDA model on diagnostic codes with 12 as the optimal number of topics and looking at the top 10 keywords in each topic along with some representative clinical notes with percent contribution of 0.80 or higher to the topic, we identified the clinical phenotypes in the form of themes in the diagnostic codes which can be abstracted as being about: (1) *Ischemic Heart Disease*, (2) *Renal Dysfunction*, (3) *Hypertensive Heart Disease*, (4) *Gastrointestinal Disease*, (5) *Skills and Self-Care Behaviors*, (6) *Cardiovascular Diseases*, (7) *Acute Heart Failure*, (8) *Heart Failure Comorbidities*, (9) *Chronic Heart Disease*, (10) *Chronic Obstructive Pulmonary Disease*, (11) *Heart Failure Disorders*, and (12) *Non-Ischemic Heart Disease*. The top 10 keywords along with their probabilities within each topic are shown in Table 1.

The theme *Ischemic Heart Disease* described coronary artery disease in patients as the keyword coronary atherosclerosis along with nonspecific chest pain was present in this theme. Chest pain or angina is a symptom of coronary artery disease. The theme *Renal Dysfunction* described chronic kidney disease and renal failure which could be present in patients with heart failure. *Hypertensive Heart Disease* described the condition of hypertension which is quite common in patients with heart failure. The common gastrointestinal manifestations of heart failure could be explained by the theme *Gastrointestinal Disease* as this theme contained the keywords like abdominal pain and other digestive/abdomen signs and symptoms and esophageal disorders.

The self-management skills in HF patients which could help them to manage their disease by following a regular routine sleep pattern and limiting the amount of salt (sodium) and fluids in the diet to avoid fluid buildup were depicted in the theme *Skills and Self-Care Behaviors*. The theme *Cardiovascular Diseases* showed the different cardiovascular comorbidities and diseases which could be experienced by HF patients like cardiac dysrhythmias, coronary atherosclerosis and other heart diseases, and acute myocardial infarction.

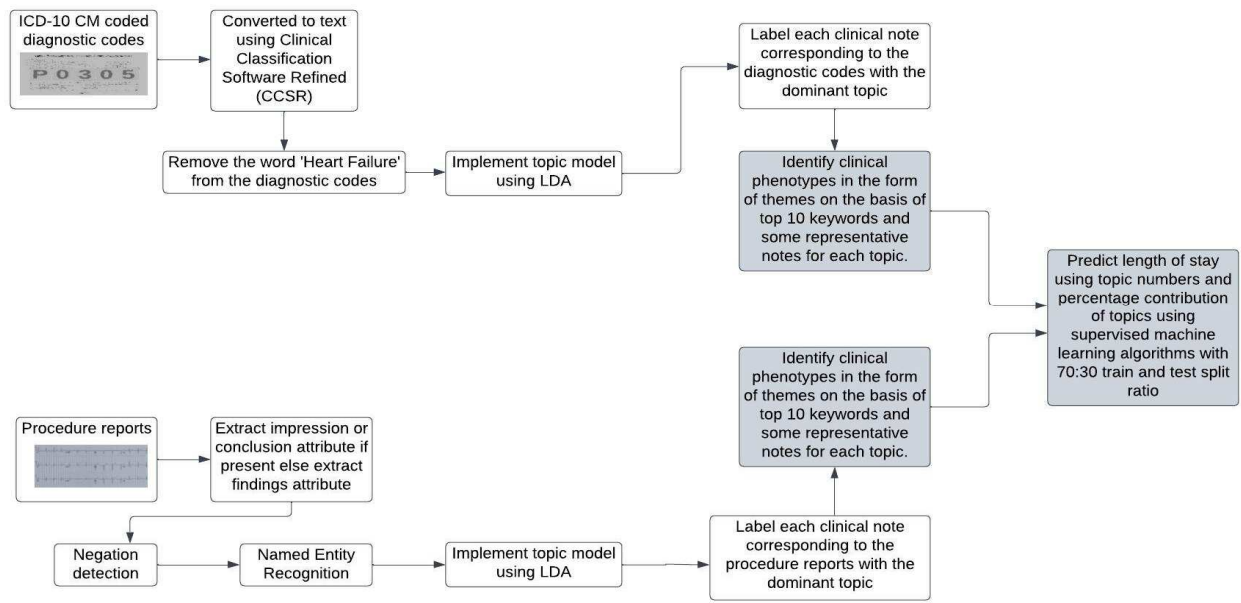

Fig. 3. Overall methodology to identify clinical phenotypes and predict length of stay in patients admitted for heart failure

A patient could experience sudden, life-threatening conditions like decompensation, common in patients experiencing acute kidney failure and cardiac arrhythmia. These conditions were explained by the theme *Acute Heart Failure,* which is evident from the keywords, acute and unspecified renal failure, acute myocardial infarction, cardiac dysrhythmias, and hypotension. Patients admitted with heart failure can have cardiovascular and non-cardiovascular comorbidities, which could affect the treatment of HF patients. So, the theme, *Heart Failure Comorbidities* was uncovered in our dataset, which contained the keywords like obesity, diabetes mellitus, Type 2, essential hypertension, chronic kidney disease, disorders of lipid metabolism, and cardiac dysrhythmias.

The theme *Chronic Heart Disease* explained the special conditions associated with chronic heart failure like chronic rheumatic heart disease and coronary atherosclerosis. Similarly, the *Chronic Obstructive Pulmonary Disease* theme explained the condition in which heart failure could cause fluid buildup in the lungs and low oxygen levels in the blood. This is one of the comorbidities associated with HF patients which affects their treatment. Some of the disorders associated with HF patients include tobacco-related disorders, sleep-wake disorders, and disorders of lipid metabolism which were explained in the theme *Heart Failure Disorders*. Finally, the last theme, *Non-Ischemic Heart Disease* described another phenotype of heart failure which includes myocarditis and cardiomyopathy.

Similarly, after implementing LDA model on procedure reports with 12 as the optimal number of topics and looking at the top 10 keywords in each topic along with some representative clinical notes with percent contribution of 0.80 or higher to the topic, we identified the themes in the procedure reports which can be abstracted as about: (1) *Ischemic Heart Disease*, (2) *Effusions*, (3) *Hypertrophic Cardiomyopathy*, (4) *Arrhythmia*, (5) *Idiopathic Cardiomyopathy*, (6) *Common Findings*, (7) *Myocardial Infarction*, (8) *Atrioventricular Septal Defect*, (9) *Congestion and Heart Block*, (10) *Cardiopulmonary Disease*, (11) *Left Ventricular Dysfunction*, and (12) *Myocardial Ischemia*. The top 10 keywords and description of each topic within each topic are shown in Table 2.

The themes in procedure reports summarized the major impressions or findings in the patients' reports. The keywords myocardial disease, atrial fibrillation, and lateral ischemia were related to the theme, *Ischemic Heart Disease.* A heart failure patient could experience both pleural and pericardial effusion, which were depicted in the theme, *Effusions*. A patient's left ventricle could be unable to pump blood in the body leading to the condition described by the theme, *Hypertrophic Cardiomyopathy*.

Similarly, there could be a problem of irregular heartbeat in the patients leading to a condition called *Arrhythmia* or sinus tachycardia most clearly identified by the ECG with a normal upright P wave in lead II preceding every QRS complex. The theme, *Idiopathic Cardiomyopathy* explained the condition found in patients with abnormalities of ventricular wall thickness, size of ventricular cavity, contraction, relaxation, conduction, and rhythm as evident from the keywords in this theme like LV dysfunction, T-wave abnormality, and occasional premature ventricular complexes.

The theme *Common Findings* summarized the commonly known abnormalities in heart failure patients. Some of these were atrial enlargement, myocardial disease, atrial fibrillation, pulmonary edema, and T-wave abnormality. The theme, *Myocardial Infarction* or a heart attack showed the condition in patients when there is a blockage of blood flow to the heart muscle. Similarly, the theme, *Atrioventricular Septal Defect*, explained the situation when there are holes between the chambers of the right and left sides of the heart and the valves are not formed correctly.

*Congestion and Heart Block* explained the vascular and venous congestion as well as the condition of heart block caused when the heartbeat is partially or completely blocked. The theme, *Cardiopulmonary Disease* explained the presence or absence of pleural edema, pleural effusion, or pneumothorax in HF patients. It included a range of serious disorders which affect both heart and lungs.

TABLE I. CLINICAL PHENOTYPES IDENTIFIED IN THE FORM OF THEMES IN THE DIAGNOSTIC CODES ALONG WITH THE TOP 10 KEYWORDS AND THEIR PROBABILITIES

| Topic Number | Theme | Keywords and their probabilities (in brackets) |
|---|---|---|
| 1. | Ischemic Heart Disease | respiratory signs and symptoms (0.027), nonspecific chest pain (0.019), cardiac dysrhythmias (0.018), fluid and electrolyte disorders (0.016), respiratory failure; insufficiency; arrest (0.015), chronic kidney disease (0.015), coronary atherosclerosis and other heart disease (0.014), aplastic anemia (0.014), acute and unspecified renal failure (0.014), nutritional anemia (0.013) |
| 2. | Renal Dysfunction | chronic kidney disease (0.017), acute and unspecified renal failure (0.017), diabetes mellitus, type 2 (0.016), personal/family history of disease (0.015), fluid and electrolyte disorders (0.015), aplastic anemia (0.015), disorders of lipid metabolism (0.013), implant, device or graft related encounter (0.013), bacterial infections (0.013), chronic obstructive pulmonary disease and bronchiectasis (0.013) |
| 3. | Hypertensive Heart Disease | chronic obstructive pulmonary disease and bronchiectasis (0.022), other general signs and symptoms (0.016), nervous system signs and symptoms (0.016), essential hypertension (0.015), obesity (0.015), cardiac dysrhythmias (0.014), respiratory failure; insufficiency; arrest (0.014), implant, device or graft related encounter (0.014), respiratory signs and symptoms (0.014), aplastic anemia (0.013) |
| 4. | Gastrointestinal Disease | chronic obstructive pulmonary disease and bronchiectasis (0.021), abdominal pain and other digestive/abdomen signs and symptoms (0.016), coronary atherosclerosis and other heart disease (0.016), esophageal disorders (0.016), implant, device or graft related encounter (0.015), abnormal findings without diagnosis (0.015), aplastic anemia (0.014), cardiac dysrhythmias (0.014), chronic kidney disease (0.014), fluid and electrolyte disorders (0.013) |
| 5. | Skills and Self-Care Behaviors | chronic kidney disease (0.019), sleep wake disorders (0.018), cardiac dysrhythmias (0.017), fluid and electrolyte disorders (0.017), obesity (0.016), other general signs and symptoms (0.016), respiratory signs and symptoms (0.016), respiratory failure; insufficiency; arrest (0.014), personal/family history of disease (0.014), disorders of lipid metabolism (0.014) |
| 6. | Cardiovascular Diseases | cardiac dysrhythmias (0.026), epilepsy; convulsions (0.019), coronary atherosclerosis and other heart disease (0.019), aplastic anemia (0.017), implant, device or graft related encounter (0.017), acute myocardial infarction (0.017), sleep wake disorders (0.017), abdominal pain and other digestive/abdomen signs and symptoms (0.016), abnormal findings without diagnosis (0.016), essential hypertension (0.016) |
| 7. | Acute Heart Failure | hypotension (0.021), acute myocardial infarction (0.017), cardiac dysrhythmias (0.016), implant, device or graft related encounter (0.014), fluid and electrolyte disorders (0.014), disorders of lipid metabolism (0.014), tobacco-related disorders (0.014), coronary atherosclerosis and other heart disease (0.014), acute and unspecified renal failure (0.014), aplastic anemia (0.013) |
| 8. | Heart Failure Comorbidities | respiratory signs and symptoms (0.027), obesity (0.023), diabetes mellitus, type 2 (0.021), essential hypertension (0.017), chronic kidney disease (0.017), disorders of lipid metabolism (0.017), asthma (0.016), cardiac dysrhythmias (0.016), acute and unspecified renal failure (0.015), coronary atherosclerosis and other heart disease (0.014) |
| 9. | Chronic Heart Disease | chronic kidney disease (0.024), implant, device or graft related encounter (0.020), fluid and electrolyte disorders (0.016), abnormal findings without diagnosis (0.016), coronary atherosclerosis and other heart disease (0.015), disorders of lipid metabolism (0.014), respiratory signs and symptoms (0.014), acute and unspecified renal failure (0.014), respiratory failure; insufficiency; arrest (0.014), chronic rheumatic heart disease (0.013) |
| 10. | Chronic Obstructive Pulmonary Disease | coronary atherosclerosis and other heart disease (0.021), acute and unspecified renal failure (0.018), chronic kidney disease (0.016), abnormal findings without diagnosis (0.016), acquired absence of limb or organ (0.015), obesity (0.015), pulmonary heart disease (0.014), pneumonia (except that caused by tuberculosis) (0.014), organ transplant status (0.013), nonrheumatic and unspecified valve disorders (0.013) |
| 11. | Heart Failure Disorders | chronic obstructive pulmonary disease and bronchiectasis (0.018), cardiac dysrhythmias (0.017), tobacco-related disorders (0.017), sleep wake disorders (0.017), obesity (0.016), coronary atherosclerosis and other heart disease (0.015), fluid and electrolyte disorders (0.015), pulmonary heart disease (0.014), chronic kidney disease (0.014), disorders of lipid metabolism (0.013) |
| 12. | Non-Ischemic Heart Disease | respiratory signs and symptoms (0.017), abnormal findings without diagnosis (0.016), chronic kidney disease (0.015), esophageal disorders (0.014), tobacco-related disorders (0.014), implant, device or graft related encounter (0.014), respiratory failure; insufficiency; arrest (0.014), personal/family history of disease (0.013), myocarditis and cardiomyopathy (0.013), acquired absence of limb or organ (0.013) |

*Left Ventricular Dysfunction* explained both systolic and diastolic conditions in heart failure patients, described as Heart failure with reduced ejection fraction (HFrEF) and Heart failure with preserved ejection fraction (HFpEF), respectively. Finally, the theme, *Myocardial Ischemia* described a common condition in HF patients where the blood flow to the heart muscle is obstructed due to the formation of plaque in the coronary artery as explained by the keywords in this theme like inferior ischemia, myocardial disease, and atrial fibrillation.

*B. Length of Stay Prediction*

The results of predicting length of stay using different machine learning algorithms and fitting topic models on diagnostic codes and procedure reports using either TF-IDF or BOW feature representations are shown in Table 3. Our results indicate that both diagnostic codes and procedure reports together improved the prediction of length of stay of patients. K-nearest neighbors achieved a maximum accuracy of 61.1% in predicting length of stay of patients on test data with results of topic models fitted on TF-IDF representation of texts. This indicates that the profiles of heart failure patients based on their length of stay can be better separated using a distance-based metric as compared to finding a hyperplane or designing rule-based algorithms.

V. DISCUSSION

The diagnostic codes and procedure reports in the clinical notes of patients can reveal different phenotypes of the underlying problem such as heart failure-related in the form of themes. By using a probabilistic framework like topic modeling, the latent structure in the typical phenotypes associated with heart failure can be revealed. These latent structures can be used to group patients based on similar disease patterns that can commonly co-occur. Additionally, a trend analysis of these phenotypes could reveal the progression of the disease in a patient.

TABLE II. CLINICAL PHENOTYPES IDENTIFIED IN THE FORM OF THEMES IN THE PROCEDURE REPORTS ALONG WITH THE TOP 10 KEYWORDS AND THEIR PROBABILITIES

| Topic Number | Theme | Keywords and their probabilities (in brackets) |
|---|---|---|
| 1. | Ischemic Heart Disease | myocardial disease (0.016), atrial fibrillation (0.010), mild cardiomegaly (0.009), abnormal ecg (0.008), pleural effusion (0.006), lateral ischemia (0.006), pulmonary congestion (0.006), t wave abnormality (0.006), left atrial enlargement (0.006), no pulmonary edema (0.005) |
| 2. | Effusions | myocardial disease (0.007), pleural effusion (0.005), abnormal ecg (0.005), no pneumothorax (0.004), pericardial effusion (0.004), platelike atelectasis (0.004), left bundle branch block (0.004), pulmonary edema (0.004), concentric left ventricular hypertrophy (0.004), lv dysfunction (0.004) |
| 3. | Hypertrophic Cardiomyopathy | pleural effusion (0.008), sinus tachycardia (0.007), mild concentric left ventricular hypertrophy (0.007), abnormal ecg (0.007), lateral ischemia (0.007), interstitial opacities (0.006), concentric left ventricular hypertrophy (0.006), t wave abnormality (0.005), inferior infarct (0.005), atrial fibrillation (0.005) |
| 4. | Arrhythmia | premature ventricular complexes (0.011), pulmonary edema (0.007), right bundle branch block (0.007), abnormal ecg (0.006), left ventricular hypertrophy (0.005), sinus tachycardia (0.005), septal infarct (0.005), atrial fibrillation (0.005), myocardial disease (0.005), pleural effusion (0.005) |
| 5. | Idiopathic Cardiomyopathy | mild concentric left ventricular hypertrophy (0.010), lv dysfunction (0.008), t wave abnormality (0.008), occasional premature ventricular complexes (0.007), abnormal ecg (0.007), left atrial enlargement (0.006), mild cardiomegaly (0.006), myocardial disease (0.006), no pneumothorax (0.006), inferior infarct (0.006) |
| 6. | Common Findings | no pleural effusion (0.010), left atrial enlargement (0.010), myocardial disease (0.009), no pneumothorax (0.009), atrial fibrillation (0.008), premature ventricular complexes (0.008), mild concentric left ventricular hypertrophy (0.007), pulmonary edema (0.007), lv dysfunction (0.007), t wave abnormality (0.006) |
| 7. | Myocardial Infarction | sinus tachycardia (0.009), no pleural effusion (0.007), no pneumonia (0.007), inferior infarct (0.006), septal infarct (0.005), myocardial disease (0.005), left atrial enlargement (0.005), mild cardiomegaly (0.005), no pulmonary edema (0.004), no pneumothorax (0.004) |
| 8. | Atrioventricular Septal Defect | myocardial disease (0.007), mild concentric left ventricular hypertrophy (0.006), left atrial enlargement (0.006), t wave abnormality (0.005), pleural effusion (0.005), septal infarct (0.004), left ventricular hypertrophy (0.004), mild cardiomegaly (0.004), right bundle branch block (0.004), atrial fibrillation (0.004) |
| 9. | Congestion and Heart Block | left atrial enlargement (0.014), sinus tachycardia (0.006), concentric left ventricular hypertrophy (0.005), pleural effusion (0.005), pulmonary vascular congestion (0.005), right bundle branch block (0.005), myocardial disease (0.004), venous congestion (0.004), moderate cardiomegaly (0.004), no pneumothorax (0.004) |
| 10. | Cardiopulmonary Disease | no pulmonary edema (0.009), no pleural effusion (0.008), no pneumothorax (0.008), mild cardiomegaly (0.007), abnormal ecg (0.007), inferior infarct (0.006), left atrial enlargement (0.006), myocardial disease (0.006), pleural effusion (0.005), t wave abnormality (0.005) |
| 11. | Left Ventricular Dysfunction | pleural effusion (0.009), lv dysfunction (0.009), mild concentric left ventricular hypertrophy (0.007), no pneumothorax (0.007), abnormal ecg (0.007), pulmonary edema (0.006), right bundle branch block (0.006), mild cardiomegaly (0.006), premature ventricular complexes (0.006), myocardial disease (0.005) |
| 12. | Myocardial Ischemia | no pleural effusion (0.012), no pneumothorax (0.010), mild cardiomegaly (0.006), septal infarct (0.005), sinus tachycardia (0.005), left atrial enlargement (0.005), pulmonary edema (0.004), inferior ischemia (0.004), myocardial disease (0.004), atrial fibrillation (0.004) |

By examining the topic modeling results for the diagnostic and procedural codes separately, we were able to glean out different aspects or similarities in HF patients in our cohort.

The results of topic modeling on diagnostic codes depicted the various paradigms of heart failure, like comorbidities, signs and symptoms, and disorders related to heart failure. It revealed different phenotypes of heart failure in the form of themes like *Ischemic* and *Non-Ischemic Heart disease*, *Hypertensive Heart Disease*, *Chronic Heart Disease*, and *Acute Heart Failure*. Additionally, the theme like *Skills and Self-Care Behaviors* contained keywords including respiratory signs and symptoms, sleep-wake disorders, obesity, personal/family history of disease. In this particular topic, the keywords were related to the non-pharmacological management of heart failure which comprises a part of successful HF treatment.

Similarly, the topic modeling results of procedure reports revealed the disease entities and specific conditions which are present in patients admitted for heart failure. Some of these entities are typically found in electrocardiogram reports, like T-wave abnormality, left axis deviation, sinus tachycardia, and abnormal ECG. Similarly, topic modeling revealed some entities commonly found in chest X-ray reports, like cardiomegaly, pulmonary edema, and left ventricular cavity size.

Longitudinal studies using these results can reveal how the phenotypic characteristics of a patient remain same or varies with time during each hospitalization. For a single patient, themes within the diagnostic codes and procedure reports can be analyzed to study the progression of a disease from one phenotypic characteristic to another. For example, for a single patient, our analysis showed that over time, a patient's disease could progress from *Chronic Heart Disease* to *Acute Heart Failure*. Additionally, for the same patient, phenotypic analysis of the procedure reports in the patient revealed that during first hospitalization, a patient belonged to a group of patients with *Left Ventricular Dysfunction* but eventually their disease progressed towards *Myocardial Ischemia*. This kind of observation coincides with the real clinical instances where left ventricular dysfunction is known to contribute towards myocardial infarction which is an acute heart disease [30]. In between the first and last hospitalization, that patient's diagnostic codes were characterized by themes like *Heart Failure Comorbidities* and *Renal Dysfunction* symbolizing that the presence of comorbidities associated with heart failure contributed towards the progression of the disease in that patient [31].

A similar observation was made in our dataset on a patient with phenotype *Heart Failure Comorbidities* at initial hospitalization who was eventually identified as having *Hypertensive Heart Disease* during next hospitalization. This patient's procedure reports revealed that the general *Common Findings* theme in the procedure reports at initial hospitalization was specifically linked to *Congestion and Heart Block* in subsequent visits. This kind of trend analysis in the phenotypes of a patient can help clinicians to monitor a progression or regression of a patient's disease. This can also help them in assisting a treatment plan for new patients by analyzing the medications administered to similar groups of patients and whether those medications helped in progression or regression of a patient's disease over time. So, this analysis emphasized the importance of considering the phenotypes identified in diagnostic codes and procedure reports together for designing a treatment plan of a patient.

TABLE III. RESULTS OF PREDICTING LENGTH OF STAY ON TEST DATA USING DIFFERENT SUPERVISED MACHINE LEARNING METHODS

| Features | Metrics | kNN | MLR | SVM | AC | RF | kNN | MLR | SVM | AC | RF |
|---|---|---|---|---|---|---|---|---|---|---|---|
| | | TF-IDF | | | | | BOW | | | | |
| Diagnostic codes topics + Percentage contribution | Precision | 0.404 | 0.110 | 0.343 | 0.333 | 0.342 | 0.374 | 0.090 | 0.303 | 0.313 | 0.311 |
| | Recall | 0.396 | 0.190 | 0.351 | 0.349 | 0.351 | 0.366 | 0.170 | 0.311 | 0.329 | 0.323 |
| | Accuracy | 0.392 | 0.183 | 0.348 | 0.342 | 0.353 | 0.362 | 0.163 | 0.308 | 0.322 | 0.322 |
| | ROC AUC | 0.672 | 0.551 | 0.653 | 0.642 | 0.641 | 0.642 | 0.521 | 0.613 | 0.622 | 0.610 |
| Procedure reports topics + Percentage contribution | Precision | 0.479 | 0.307 | 0.415 | 0.384 | 0.364 | 0.449 | 0.297 | 0.375 | 0.364 | 0.333 |
| | Recall | 0.463 | 0.342 | 0.401 | 0.377 | 0.354 | 0.433 | 0.322 | 0.361 | 0.357 | 0.328 |
| | Accuracy | 0.459 | 0.329 | 0.396 | 0.369 | 0.378 | 0.429 | 0.309 | 0.356 | 0.349 | 0.342 |
| | ROC AUC | 0.730 | 0.682 | 0.701 | 0.673 | 0.681 | 0.698 | 0.662 | 0.661 | 0.653 | 0.651 |
| Both diagnostic and procedure reports topics and Percentage contribution | Precision | **0.605** | 0.284 | 0.424 | 0.366 | 0.378 | 0.575 | 0.264 | 0.384 | 0.346 | 0.348 |
| | Recall | **0.611** | 0.354 | 0.429 | 0.381 | 0.394 | 0.581 | 0.334 | 0.389 | 0.361 | 0.364 |
| | Accuracy | **0.611** | 0.343 | 0.423 | 0.374 | 0.398 | 0.581 | 0.323 | 0.383 | 0.354 | 0.368 |
| | ROC AUC | **0.828** | 0.689 | 0.723 | 0.681 | 0.711 | 0.798 | 0.669 | 0.683 | 0.661 | 0.681 |

These phenotypic characteristics in a patient can also be used to predict length of stay in a patient which could help health professionals to manage the available resources for any current or new hospitalizations [32]. Our results indicate that both diagnostic codes and procedure reports together improved the prediction of length of stay of patients. K-nearest neighbors achieved a maximum accuracy of 61.1% and ROC AUC of 0.828 in predicting length of stay of patients which indicates that profiles of heart failure patients can be better separated using a distance-based metric as compared to finding a hyperplane or designing rule-based algorithm. Additionally, topic modeling results using only procedure reports performed better as compared to results using only diagnostic codes to predict length of stay of patients which indicates that procedure reports are a better indicator of predicting a patient's stay.

As procedure reports of a patient vary with time on the basis of whether a patient's condition is improving or worsening but diagnostic codes remain fairly the same during a particular hospitalization of a patient, we found that the former were better at predicting the length of stay of a patient. Although there was an improvement in the accuracy when results from both diagnostic codes and procedure reports were combined.

This framework of analysis could promote digital health and can be generalizable across different hospital settings and coding standards in diagnostic and procedure reports. As this is the first study in predicting length of stay in patients admitted for heart failure using clinical notes, an accuracy of 61.1% and ROC AUC of 0.828 showed that studying the patterns in the diagnostic codes and procedure reports of patients in the form of phenotypes played an important role in predicting length of stay in heart failure patients.

This study has its own limitations. As, we used the dataset from a single hospital, the records for all the patients during each stay was maintained consistently electronically. But, while analyzing the clinical notes of other hospitals, there could be some instances where some notes are hand-written and are missing. So, we need to further implement our approach on such clinical notes provided by other hospitals and compare the results. Further, this study focuses on studying the phenotypic characteristics of patients admitted for heart failure by relying on tokenizing the content in the procedure and diagnostic codes to train our machine learning models. For comparative purposes, we plan to extend this work by implementing large language models on these text-based data and train deep learning models to predict length of stay of patients.

In the future work, physiological vitals, medications, and laboratory test results can also be incorporated to study their influence on the phenotypic characteristics of a patient and to train machine learning models to predict length of stay with a higher accuracy. Finally, we would need to consider the demographic features of a patient like age, gender, ethnicity, and insurance status to study the fairness and bias in our machine learning models in predicting length of stay of the patients.

## VI. CONCLUSION

Heart failure is a disorder and its management in the patients involves understanding its different phenotypes which usually have a probabilistic structure. Analysis of clinical notes of patients through LDA using the data about diagnostic codes and procedure reports during each hospitalization of a patient revealed different phenotypes associated with heart failure in the form of themes. These themes can be further used to predict length of stay in these patients with an accuracy of 61.1% using K-nearest neighbors.